\documentclass[letterpaper]{article} 
\usepackage{aaai2026}  
\usepackage{times}  
\usepackage{helvet}  
\usepackage{courier}  
\usepackage[hyphens]{url}  
\usepackage{graphicx} 
\usepackage{natbib}  
\usepackage{caption} 
\usepackage{algorithm}
\usepackage{algorithmic}

\usepackage{newfloat}
\usepackage{listings}
\DeclareCaptionStyle{ruled}{labelfont=normalfont,labelsep=colon,strut=off} 
\floatstyle{ruled}
\newfloat{listing}{tb}{lst}{}
\floatname{listing}{Listing}
\title{IGFuse: Interactive 3D Gaussian Scene Reconstruction via Multi-Scans Fusion}

\author {
    Wenhao Hu\textsuperscript{\rm 1,4}$^{*}$,
    Zesheng Li\textsuperscript{\rm 3},
    Haonan Zhou\textsuperscript{\rm 2},
    Liu Liu\textsuperscript{\rm 4},
    Xuexiang Wen\textsuperscript{\rm 2},
    Zhizhong Su\textsuperscript{\rm 4}, \newline
    Xi Li\textsuperscript{\rm 1},
    Gaoang Wang\textsuperscript{\rm 1,2}$^{\dag}$
}
\affiliations {
    \textsuperscript{\rm 1}College of Computer Science and Technology, Zhejiang University\\
    \textsuperscript{\rm 2}ZJU-UIUC Institute, Zhejiang University\\
    \textsuperscript{\rm 3}Nanyang Technological University\\
    \textsuperscript{\rm 4}Horizon Robotics\\
}

\usepackage{amsmath} 
\usepackage{booktabs} 
\usepackage{multirow}

\usepackage{graphicx} 
\usepackage{bbding}
\usepackage{makecell}
\usepackage{cuted}
\begin{document}

\nocopyright
\maketitle
\begingroup
\renewcommand\thefootnote{*}
\footnotetext{This work was done during an internship at Horizon Robotics.}
\renewcommand\thefootnote{\dag}
\footnotetext{Corresponding author.}
\endgroup
\vspace{-20em}

\begin{strip}
\centering
\vspace*{-16mm}
\includegraphics[width=1\textwidth]{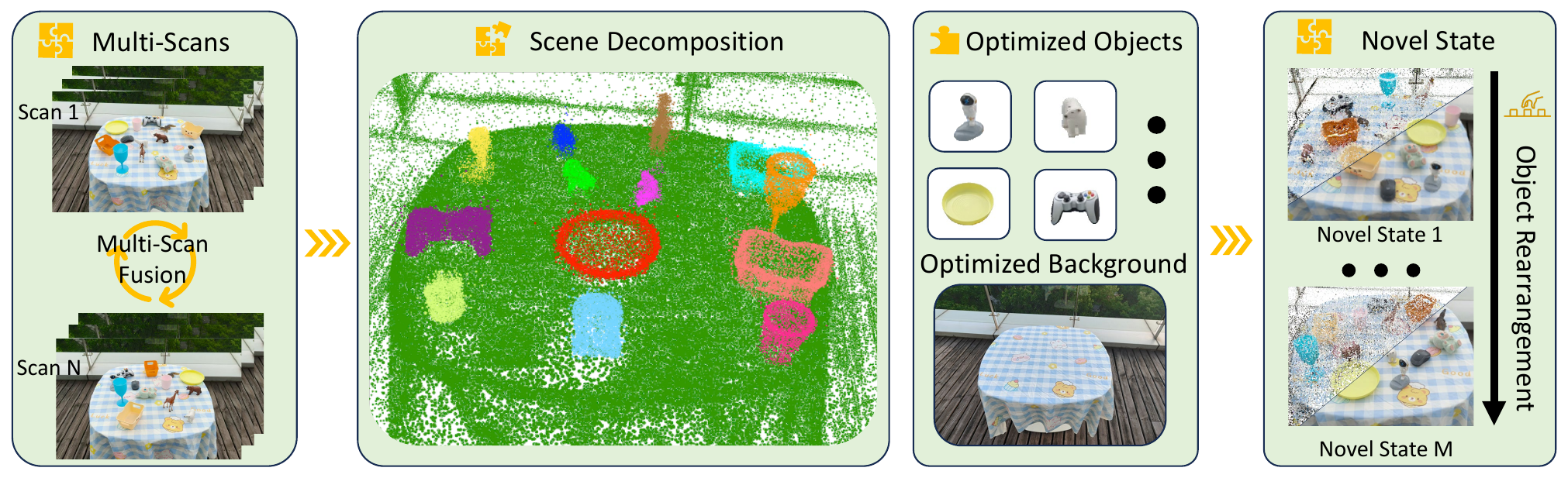} 
\captionof{figure}{Given multiple observed scene scans, we perform multi-state optimization to jointly reconstruct consistent Gaussian fields. The scene is then decomposed into objects and background, which are jointly represented and constrained across scans. This enables the interactive generation of new scene states with coherent object compositions and realistic rendering.}
\label{fig:stripimage}
\end{strip}

\begin{abstract}
Reconstructing complete and interactive 3D scenes remains a fundamental challenge in computer vision and robotics, particularly due to persistent object occlusions and limited sensor coverage. Multi-view observations from a single scene scan often fail to capture the full structural details. Existing approaches typically rely on multi-stage pipelines—such as segmentation, background completion, and inpainting—or require per-object dense scanning, both of which are error-prone, and not easily scalable.
We propose \textbf{IGFuse}, a novel framework that reconstructs interactive Gaussian scene by fusing observations from multiple scans, where natural object rearrangement between captures reveal previously occluded regions. Our method constructs segmentation-aware Gaussian fields and enforces bi-directional photometric and semantic consistency across scans. To handle spatial misalignments, we introduce a pseudo-intermediate scene state for unified alignment, alongside collaborative co-pruning strategies to refine geometry.
IGFuse enables high-fidelity rendering and object-level scene manipulation without dense observations or complex pipelines. Extensive experiments validate the framework’s strong generalization to novel  scene configurations, demonstrating its effectiveness for real-world 3D reconstruction and real-to-simulation transfer. Our project page is available at \url{https://whhu7.github.io/IGFuse}
\end{abstract}

\section{Introduction}
\label{intro}
\begin{figure*}[t]
    \centering
\includegraphics[width=\linewidth]{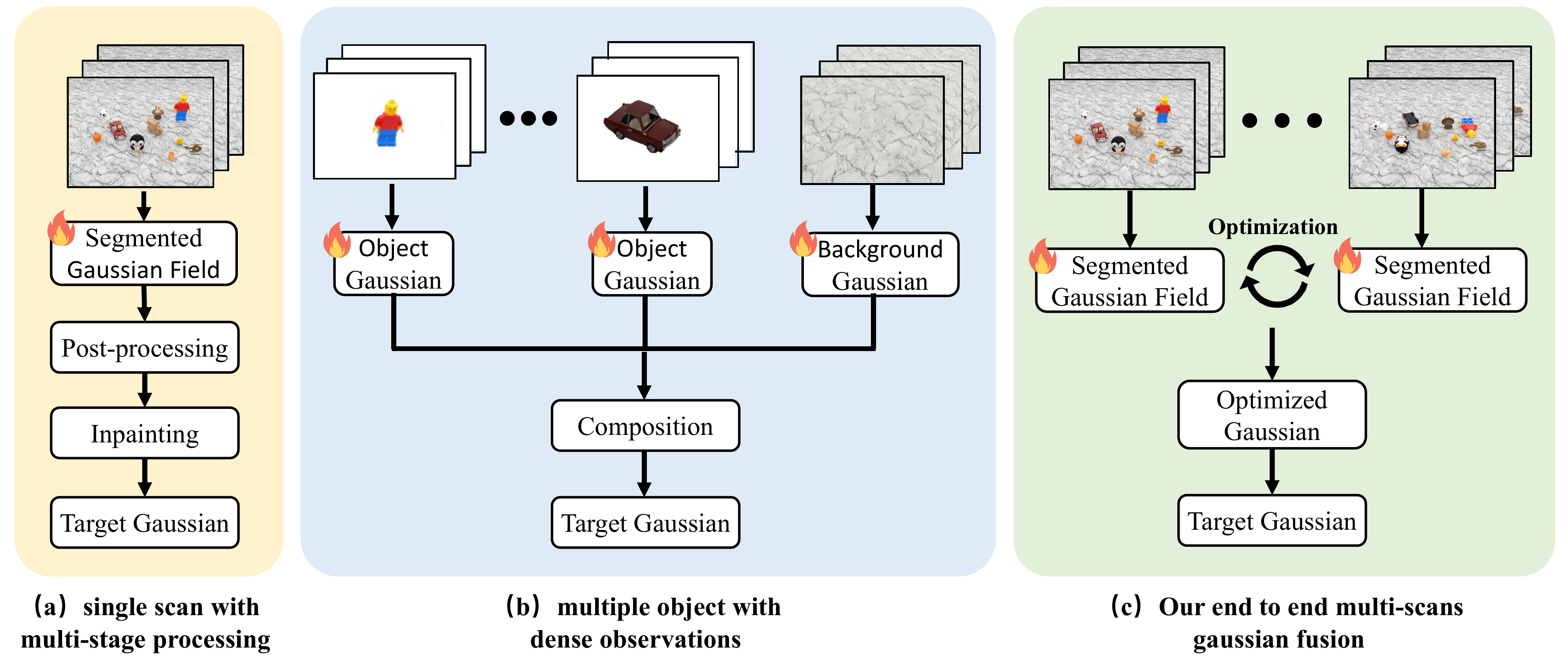}
  \caption{Comparison of different paradigms for constructing interactive 3D Gaussian. 
(a) Traditional single-scan pipelines rely on multi-stage post-processing and inpainting, which may introduce accumulated artifacts. 
(b) Object-centric approaches require dense multi-view observations of all components, followed by explicit composition. 
(c) Our proposed end-to-end multi-scans fusion model jointly optimizes multi-state Gaussian fields via cross-state supervision, effectively compensating for occlusions across different observations and enabling interactive Gaussian reconstruction. }
    \label{fig:teaser}
\end{figure*}

Reconstructing interactive 3D scenes from partially observed environments remains a core challenge in vision and robotics~\cite{zhu20243d, wang2024nerf, pang2025learning, mendonca2023structured}. Recent advances in 3D Gaussian Splatting~\cite{kerbl20233d} have enabled explicit scene representations by modeling geometry and appearance using compact Gaussian primitives. Some approaches, such as Gaussian Grouping~\cite{ye2023gaussian} and DecoupledGaussian~\cite{wang2025decoupledgaussian}, aim to support interactive scene reconstruction by combining instance-level segmentation with inpainting-based refinement. While partially effective, these multi-stage pipelines face several challenges. Feature-based segmentation often produces inaccuracies—especially near object boundaries and occluded regions—resulting in misclassified Gaussians, and visual artifacts. These issues require additional post-processing, which increases system complexity. Furthermore, inpainting methods frequently fail to recover fine background details, leading to unrealistic or blurry reconstructions. These limitations compromise the overall fidelity and consistency of the reconstructed scene and reduce the system’s reliability in downstream applications involving object-level understanding or manipulation.

In parallel, recent research has explored integrating 3D Gaussian Splatting into interactive and physically grounded simulation frameworks~\cite{barcellona2024dream, yu2025real2render2real, yang2025novel, lou2024robo, han2025re, zhu2025vr}. Methods such as RoboGSim~\cite{li2024robogsim} and SplatSim~\cite{qureshi2024splatsim} leverage Gaussian representations to construct photorealistic virtual environments from real-world observations. However, these approaches typically depend on dense multi-view object captures to achieve high-fidelity reconstructions, which limits scalability in practical scenarios. 

To address these limitations, we propose leveraging multiple observations of the same scene captured under natural object rearrangements caused by human interactions. These interaction-driven scene states expose previously occluded areas and implicitly provide geometric cues for refining segmentation and structure. Motivated by these insights, we introduce \textbf{IGFuse}, a novel framework for reconstructing interactive 3D scenes by fusing observations across multiple scans. Our method constructs segmentation-aware Gaussian fields for each scan and jointly optimizes them by enforcing bi-directional photometric and semantic consistency. To align scans captured under different scene layouts, we introduce a pseudo scene state that serves as a intermediate reference frame. Additionally, we design collaborative co-pruning strategies to suppress misaligned or inconsistent Gaussians and enhance geometric completeness.

IGFuse enables high-fidelity rendering and object-level scene manipulation—without requiring dense view captures, or multi-stage pipelines. Our framework generalizes well to novel rearranged scene states, offering a scalable and robust solution for 3D scene reconstruction in interactive environments. In summary, our main contributions are:
\begin{itemize}
\item We propose \textbf{IGFuse}, a framework for interactive 3D scene reconstruction from multi-scan observations driven by real-world object rearrangements.
\item We construct segmentation-aware Gaussian fields and enforce bi-directional photometric and semantic consistency across scans to jointly complete the scene.
\item We introduce a pseudo-intermediate Gaussian state for unified alignment across perturbed scene configurations, improving fusion quality and geometric coherence.
\end{itemize}

\section{Related Works}

\subsection{3D Gaussian Segmentation} 
Recent methods have extended Gaussian Splatting to perform scene segmentation~\cite{zhu2025rethinking,hu2025pointmap,hu2024semantic}.
GaussianEditor~\cite{chen2024gaussianeditor} projects 2D segmentation masks onto 3D Gaussians via inverse rendering.
Gaussian Grouping~\cite{ye2023gaussian} attaches segmentation features to each Gaussian and aligns multi-view IDs using video segmentation~\cite{cheng2023tracking}, while Gaga~\cite{lyu2024gaga} resolves cross-view inconsistencies via a 3D-aware memory bank.
FlashSplat~\cite{shen2024flashsplat} proposes a fast, globally optimal LP-based segmentation method.
OpenGaussian~\cite{wu2024opengaussian} and InstanceGaussian~\cite{li2024instancegaussian} use contrastive learning for point-level segmentation.
GaussianCut~\cite{jain2024gaussiancut} formulates a graph-cut optimization to separate foreground and background.
COB-GS~\cite{zhang2025cob} improves boundary precision via adaptive splitting and visual refinement.

However, 3D segmentation alone is insufficient for interactive reconstruction, as 2D biases often result in flawed 3D masks. This necessitates post-processing and inpainting~\cite{liu2024infusion, cao2024mvinpainter, huang20253d} to fill gaps caused by object movement—leading to a complex and error-prone pipeline.
In contrast, our method fuses multi-scan observations under varied configurations to achieve mutual visibility and end-to-end reconstruction without explicit inpainting. Object transitions help calibrate segmentation errors, producing clean and consistent 3D Gaussians suited for interaction tasks.
\subsection{Interactive Scene Reconstruction} 
Some approaches simulate real-world interactions by constructing implicit generative models from video data. UniSim~\cite{yang2023learning} predicts visual outcomes conditioned on diverse actions using an autoregressive framework over heterogeneous datasets. iVideoGPT~\cite{wu2024ivideogpt} encodes observations, actions, and rewards into token sequences for scalable next-token prediction via compressive tokenization. However, these methods often lack 3D and physical consistency and are generally difficult to train. Recent work focuses on enabling interactive simulators by integrating reconstructed real scenes into physics engines. RoboGSim~\cite{li2024robogsim} embeds 3D Gaussians into Isaac Sim. SplatSim~\cite{qureshi2024splatsim} replaces meshes with Gaussian splats for photorealistic rendering. PhysGaussian~\cite{xie2024physgaussian} and Spring-Gaus~\cite{zhong2024reconstruction} enable mesh-free physical simulation using Newtonian or elastic models. NeuMA~\cite{cao2024neuma} refines simulation using image-space gradients. However, these methods typically rely on dense, per-object 3D capture. In contrast, our method is more lightweight and scalable—requiring only a few multi-scan observations under varying scene configurations.

\section{Method}
\label{method}

\subsection{Preliminary}
\label{pre}
Segmented Gaussian Splatting~\cite{ye2023gaussian} models a scene as a set of 3D Gaussians, each parameterized as $\mathcal{G} = \{\boldsymbol{x}, \boldsymbol{\Sigma}, \boldsymbol{\alpha}, \boldsymbol{c}, \boldsymbol{s}\}$, where $\boldsymbol{x}$ denotes the 3D center position, $\boldsymbol{\Sigma}$ represents the spatial covariance matrix, $\boldsymbol{\alpha}$ is the opacity coefficient, $\boldsymbol{c}$ is the RGB color vector, and $\boldsymbol{s}$ is a learnable feature vector used for segmentation. 

During rendering, each Gaussian is projected onto the 2D image plane using a differentiable $\alpha$-blending mechanism. Both the final pixel color $C$ and segmentation feature $S$ are computed by accumulating Gaussian contributions weighted by their projected opacities $\boldsymbol{\alpha}_i'$:
\begin{align}
C = \sum_{i \in \mathcal{N}} \boldsymbol{c}_i \boldsymbol{\alpha}_i' \prod_{j=1}^{i-1}(1 - \boldsymbol{\alpha}_j'), \quad
S = \sum_{i \in \mathcal{N}} \boldsymbol{s}_i \boldsymbol{\alpha}_i' \prod_{j=1}^{i-1}(1 - \boldsymbol{\alpha}_j')
\label{eq:rendering}
\end{align}

\subsection{Modeling from Multi-Scan Observations}
Given a set of scans ${\mathcal{X}_1, \mathcal{X}_2, \ldots, \mathcal{X}_N}$, where each scan $\mathcal{X}_i = (\mathcal{I}_i, \mathcal{S}_i)$ contains image observations $\mathcal{I}_i$ and segmentation masks $\mathcal{S}_i$ captured under different object configurations, our goal is to fuse multi-scan observations and construct an interactive 3D scene representation. This representation supports realistic rendering under interaction signals, where arbitrary object movements produce plausible and consistent results.

To achieve this, we treat each scan as a discrete scene state and construct a corresponding segmentation-aware Gaussian field $\mathcal{G}_i$, where $i \in {1, 2, \ldots, N}$. These Gaussian fields encode geometry, appearance, and segmentation under different object layouts. The differences across fields $\{\mathcal{G}_1, ..., \mathcal{G}_N\}$ reflect object-level interactions and structural changes in the scene.

To integrate information across scans, we adopt a training strategy that randomly samples a pair $(\mathcal{G}_i, \mathcal{G}_j)$ in each epoch. Using known rigid object transformations, we align the pair and fuse their information by enforcing bi-directional photometric and semantic consistency. This enables mutual supervision, helping to refine occlusion-prone regions and correct segmentation errors. The fusion process is formulated as a joint optimization:

\begin{align}
(\mathcal{G}_i^*, \mathcal{G}_j^*) = \arg\min_{\mathcal{G}_i, \mathcal{G}_j} \; \mathcal{L}_{\text{joint}}
\label{eq:optimization}
\end{align}

Given the optimized fields and transformation $T$, we synthesize a new interactive scene configuration $\mathcal{G}_t$ through explicit Gaussian transformation:

\begin{align}
\{\mathcal{G}_i^*, \mathcal{G}_j^*\} \xrightarrow{T} \mathcal{G}_t
\label{eq:state_transfer}
\end{align}

By jointly optimizing over scan pairs and explicitly modeling object-level transformation, our framework constructs a coherent and manipulable 3D Gaussian scene representation without relying on dense captures or multi-stage post-processing.

\subsection{Gaussian State Transfer}

\begin{figure*}[t]
    \centering
    \includegraphics[width=\linewidth]{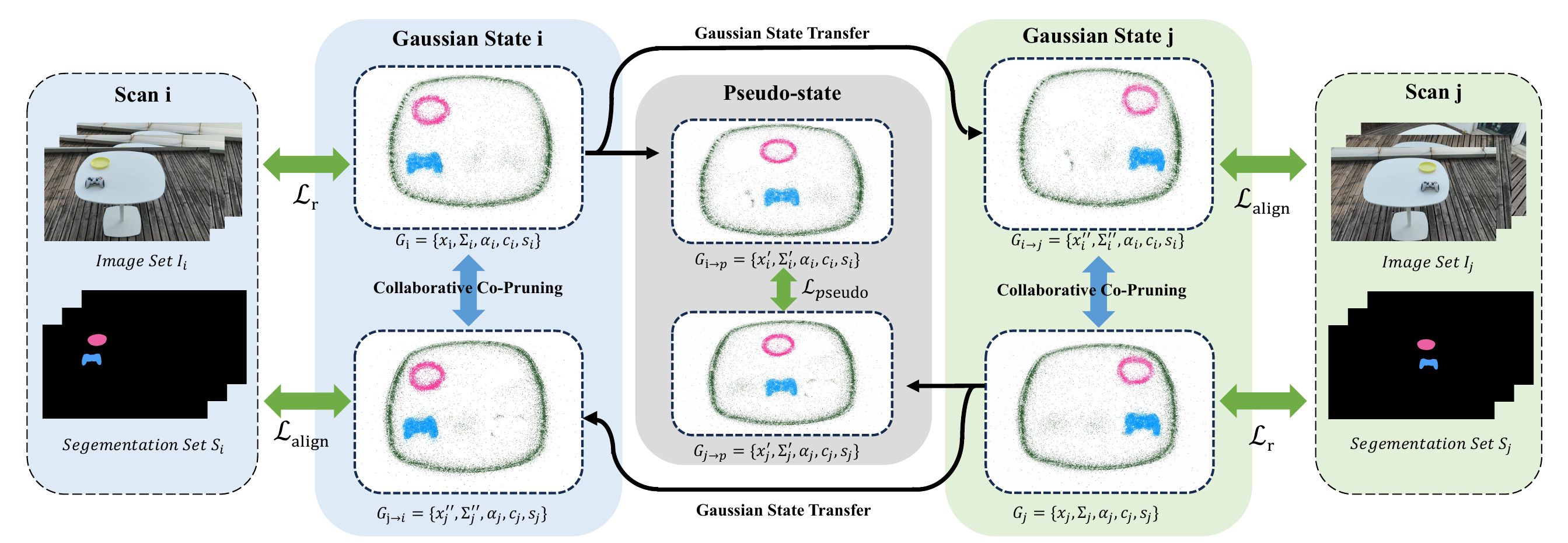}
    \caption{Overview of our dual-state Gaussian alignment pipeline. Given two input scans (scan i and scan j), the Gaussians in state i are initially constrained by corresponding image observations. After transferring to state j (i.e., $G_i \rightarrow G_{i \rightarrow j}$), the Gaussians are further supervised by state j’s image via an alignment loss $\mathcal{L}_{\text{align}}$, and regularized through a co-pruning strategy that enforces 3D consistency by removing mismatched or redundant components. The reverse transfer ($G_j \rightarrow G_{j \rightarrow i}$) is performed symmetrically. Additionally, both states are transferred into a shared pseudo-state space ($G_{i \rightarrow p}$, $G_{j \rightarrow p}$), where a pseudo-state loss $\mathcal{L}_{\text{pseudo}}$ encourages tighter cross-state alignment.}
    \label{fig:pipeline}
\end{figure*}

To model scene-level transformations, the operator $\mathbf{T}$ is defined as an object-aware function that applies per-Gaussian rigid transformations based on semantic identity. Let the Gaussian field be decomposed into foreground and background subsets:
\begin{align}
\mathcal{G} = \mathcal{G}_\text{fg} \cup \mathcal{G}_\text{bg}, \quad \mathcal{G}_\text{fg} = \bigcup_{o=1}^{O} \mathcal{G}_\text{fg}^{(o)}
\end{align}
where each foreground object $o$ is associated with a rigid transformation $\mathbf{T}^{(o)}$. For any Gaussian $\boldsymbol{g}_i \in \mathcal{G}$, let $o_i$ denote the object to which it belongs. Then, $\mathbf{T}$ is applied as:
\begin{align}
\mathbf{T}(\boldsymbol{g}_i) =
\begin{cases}
\mathbf{T}^{(o_i)} \cdot \boldsymbol{g}_i, & \text{if } \boldsymbol{g}_i \in \mathcal{G}_\text{fg} \\
\boldsymbol{g}_i, & \text{if } \boldsymbol{g}_i \in \mathcal{G}_\text{bg}
\end{cases}
\end{align}
This formulation ensures spatially consistent transformation and geometric fidelity of object-level transformation while preserving the static background.

\subsection{Bidirectional Alignment}
\label{sec:Mapping}
To ensure geometric and semantic consistency across different scene states, we enforce that the rendered outputs from transformed Gaussian fields align with the ground-truth observations in the corresponding target states. As mentioned before, we apply transformation $\mathbf{T}_{i\to j}$ to $\mathcal{G}_i$ and transformation $\mathbf{T}_{j\to i}$ to $\mathcal{G}_j$. For any viewpoint $v$, the transformed Gaussian fields are rendered into RGB images and segmentation masks, which are then compared with the corresponding ground-truth observations $(\mathcal{I}_i^v, \mathcal{S}_i^v)$ and $(\mathcal{I}_j^v, \mathcal{S}_j^v)$ from the original states. The total alignment loss combines photometric and segmentation consistency, defined as:

\begin{align}
\mathcal{L}_{\text{align}}(\mathcal{G}_i,\mathcal{G}_j,\theta) = 
&\left\| R\left( \mathbf{T}_{i\to j}(\mathcal{G}_i), v \right) - \mathcal{I}_j^v \right\|_1 \nonumber \\
&+ \left\| R\left( \mathbf{T}_{j\to i}(\mathcal{G}_j), v \right) - \mathcal{I}_i^v \right\|_1 \nonumber \\
&+ \mathrm{CE}\left( f_\theta\left(\mathcal{M}\left( \mathbf{T}_{i\to j}(\mathcal{G}_i), v \right)\right), \mathcal{S}_j^v \right) \nonumber \\
&+ \mathrm{CE}\left( f_\theta\left(\mathcal{M}\left( \mathbf{T}_{j\to i}(\mathcal{G}_j), v \right)\right), \mathcal{S}_i^v \right)
\label{eq:align_loss}
\end{align}
where $R(\cdot, v)$ and $M(\cdot, v)$ denote the rendering functions that generate the RGB image and segmentation feature from viewpoint $v$, as defined in Equation~\ref{eq:rendering}. The segmentation output is obtained via a shared classifier $f_\theta$, which is jointly applied to both $\mathcal{G}_i$ and $\mathcal{G}_j$. Specifically, $f_\theta$ consists of a linear layer that projects each identity embedding to a $(K+1)$-dimensional space, where $K$ is the number of instance masks in the 3D scene~\cite{ye2023gaussian}. The cross-entropy loss $\mathrm{CE}(\cdot, \cdot)$ measures the semantic alignment between predicted and ground-truth masks. 

This bidirectional consistency encourages each transformed Gaussian field to accurately reconstruct the scene content of the opposite state, thereby reinforcing object-level correspondence and enhancing alignment across different scene configurations.

\subsection{Pseudo-state Guided Alignment}
\label{sec:Pseudo}

To enhance the generalizability of the interactive Gaussian across diverse scene configurations, we introduce a pseudo-state $\mathcal{G}_p$ that serves as an intermediate reference for supervision. This pseudo-state is constructed by applying geometric constraints, such as collision and boundary regularization, to synthesize a virtual configuration between the two observed states. Unlike the original states, the pseudo-state is not tied to any specific observation but provides a common state that facilitates consistent alignment between $\mathcal{G}_i$ and $\mathcal{G}_j$.

We compute transformation matrices $\mathbf{T}_{i\to{p}}$ and $\mathbf{T}_{j\to{p}}$ to transfer the original fields $\mathcal{G}_i$ and $\mathcal{G}_j$ into $\mathcal{G}_p$. By transforming both fields into this shared pseudo-state, we enable direct comparison and alignment of their rendered outputs. Specifically, we render the transformed fields from the same viewpoint $v$ and enforce photometric and semantic consistency between them. The corresponding loss is defined as:

\begin{align}
\mathcal{L}_{\text{pseudo}}(\mathcal{G}_i,\mathcal{G}_j,\theta) = 
&\left\| R\left( \mathbf{T}_{i\to{p}}(\mathcal{G}_i), v \right) 
- R\left( \mathbf{T}_{j\to{p}}(\mathcal{G}_j), v \right) \right\|_1 \nonumber \\
&+ \mathrm{CE}\Bigl( f_\theta\left(\mathcal{M}\left( \mathbf{T}_{i\to{p}}(\mathcal{G}_i), v \right)\right), \nonumber \\
&\qquad f_\theta\left(\mathcal{M}\left( \mathbf{T}_{j\to{p}}(\mathcal{G}_j), v \right)\right) \Bigr)
\label{eq:pseudo_loss}
\end{align}

By leveraging a dynamically constructed pseudo-state as an adaptive supervision signal, the model can better reconcile differences between the two input states and generalize more effectively to unseen or intermediate scene configurations.

\subsection{Collaborative Co-Pruning}
\label{sec:Pruning}
Inspired by geometric consistency-based filtering strategies~\cite{zhang2024cor}, we introduce a co-pruning mechanism to suppress residual artifacts arising from imperfect segmentation during cross-state Gaussian transfer. The mechanism removes spatially inconsistent Gaussians by evaluating geometric agreement between the two states.  When a Gaussian field is transferred from one state to another, unmatched or misaligned points may remain due to occlusion, noise, or over-segmentation. Our strategy prunes these outliers by checking whether transferred Gaussians can be reliably explained by the geometry of the target field.

For each transformed Gaussian $\boldsymbol{g}_k \in \mathbf{T}_{i\to j}(\mathcal{G}_i)$, we identify its nearest neighbor $\boldsymbol{g}_l \in \mathcal{G}_j$ using Euclidean distance. A Gaussian is marked for pruning if the spatial deviation between $\boldsymbol{g}_k$ and $\boldsymbol{g}_l$ exceeds a predefined threshold $\tau$. The binary pruning indicator $m_i$ is computed as:
\begin{align}
m_i =  \mathbf{1}\left( \left\| \boldsymbol{x}_k - \boldsymbol{x}_l \right\|_2 > \tau \right)
\end{align}
where $\boldsymbol{x}_k$ and $\boldsymbol{x}_l$ are the 3D centers of $\boldsymbol{g}_k$ and $\boldsymbol{g}_l$, and $\mathbf{1}(\cdot)$ denotes the indicator function. Gaussians with $m_i = 1$ are discarded as unreliable or redundant. A symmetric process is applied in the opposite direction, using $\mathcal{G}_j$ transformed to the frame of $\mathcal{G}_i$ to prune outliers in $\mathcal{G}_j$, resulting in a collaborative co-pruning scheme.

\subsection{Training Objective}
The overall training objective combines three loss terms:

\begin{align}
\mathcal{L}_{joint}(\mathcal{G}_i,\mathcal{G}_j,\theta) 
&= \mathcal{L}_{\text{r}}(\mathcal{G}_i,\theta) + \mathcal{L}_{\text{r}}(\mathcal{G}_j,\theta)+ \lambda_a \mathcal{L}_{\text{align}}(\mathcal{G}_i,\mathcal{G}_j,\theta) \nonumber \\
&\quad  +\lambda_p \mathcal{L}_{\text{pseudo}}(\mathcal{G}_i,\mathcal{G}_j,\theta)
\end{align}
where $\mathcal{L}_{\text{r}}$ denotes the same reconstruction loss adopted from Gaussian Grouping~\cite{ye2023gaussian} (detailed in the appendix), $\mathcal{L}_{\text{align}}$ enforces bidirectional rendering consistency, and $\mathcal{L}_{\text{pseudo}}$ introduces regularization through pseudo-state supervision. The weights $\lambda_a$ and $\lambda_p$ are used to balance the contributions of each term.

\section{Experiment}
\label{exp}
\subsection{Dataset}
\label{dataset}
To support multi-scan scene modeling, we construct both synthetic and real-world datasets. The synthetic dataset is generated in Blender~\cite{blender}, where $N$ textured objects from BlenderKit~\cite{blenderkit} are placed within a static background. Additional scans are created by randomly altering object poses to reflect different interaction states. Real-world data is captured in a similar manner using handheld RGB cameras, resulting in 7 synthetic and 5 real scenes. For evaluation, we generate a test configuration for each scene by randomly repositioning objects. We then render images from predefined camera views and compute PSNR and SSIM against ground truth images to assess interaction fidelity under novel object arrangements. Further implementation and dataset details are provided in the appendix.
\subsection{Experimental Setup}
\paragraph{Implementation details}
During training, we first optimize the segmented Gaussians using only $\mathcal{L}_{\text{r}}$ for 10,000 epochs, then jointly train with $\mathcal{L}_{\text{align}}$ and $\mathcal{L}_{\text{pseudo}}$ to refine the dual Gaussian field for another n$\times$5000 epochs, where n is the total number of scans in the scene.
The output classification linear layer has 16 input channels and 256 output channels. 
The pruning threshold parameter $\tau$ is set to 0.5. In training, we set $\lambda_a=1.0$  and $\lambda_p=1.0$. We use the Adam optimizer for both gaussians and linear layer, with a learning rate of 0.0025 for segmentation feature and 0.0005 for linear layer. All datasets are trained  on a single NVIDIA 4090 GPU.

\paragraph{Baselines}
We compare our method with representative Gaussian Splatting-based scene modeling frameworks. Existing pipelines often involve multi-stage processing, including segmentation, background completion, inpainting, and fine-tuning. We include several representative segmentation methods in our comparison. GaussianEditor~\cite{chen2024gaussianeditor} performs segmentation through inverse rendering optimization. Gaussian Grouping~\cite{ye2023gaussian} clusters Gaussians based on feature similarity. GaussianCut~\cite{jain2024gaussiancut} formulates segmentation as a graph-cut optimization problem over Gaussian primitives. We also include Decoupled Gaussian~\cite{wang2025decoupledgaussian}, which segments objects using Gaussian segmentation feature, then performs remeshing and LaMa-based refinement to complete the scene.

\subsection{Novel State Synthesis}
\begin{figure*}[!ht]
    \centering
    \includegraphics[width=1\linewidth]{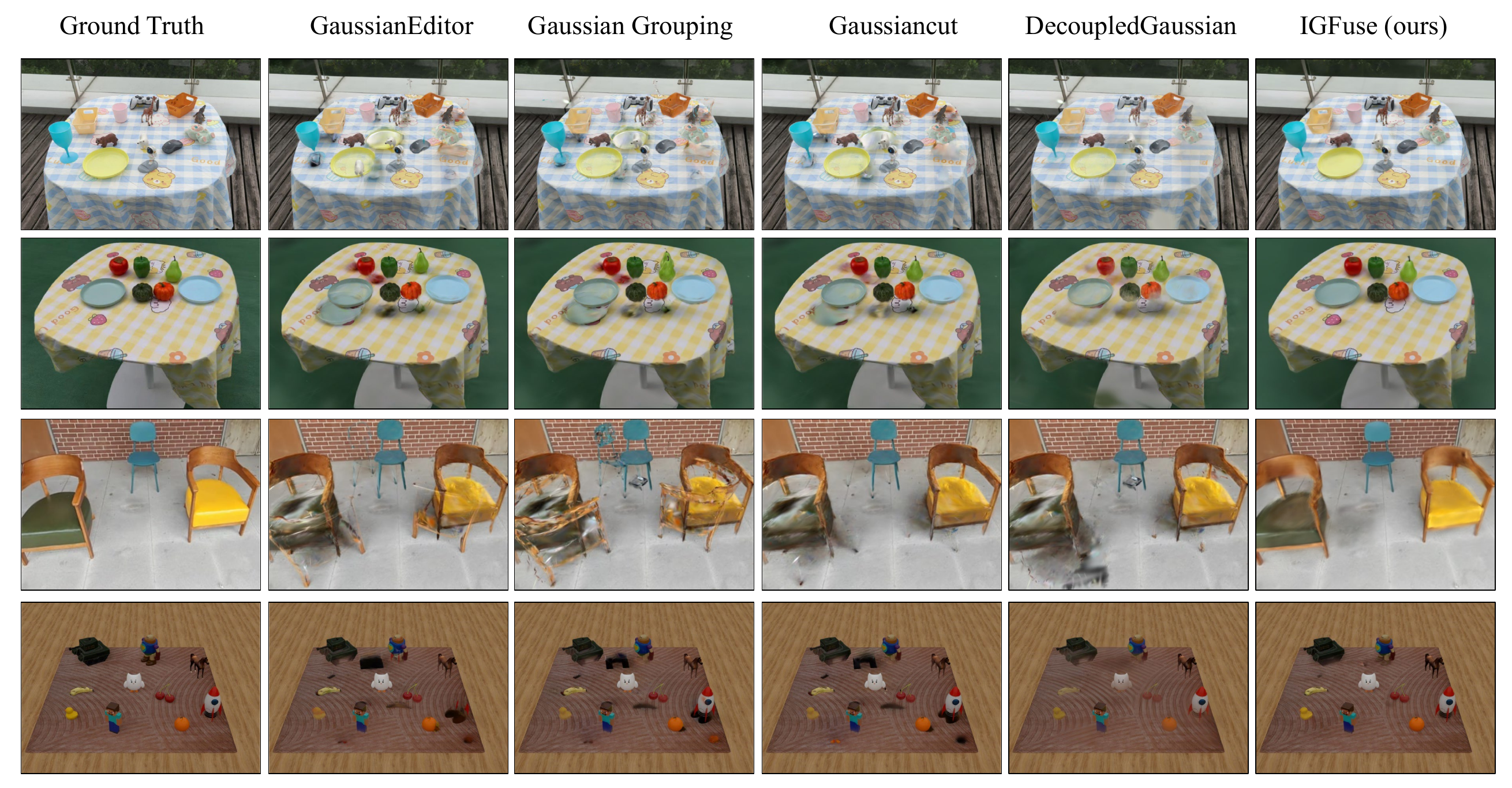}
    \caption{Qualitative comparison of novel state synthesis under different pipelines. We evaluate on both real-world scenes (top three) and a synthetic scene. While existing methods struggle with object mixing, boundary artifacts, or background corruption, our method achieves significantly more accurate and complete novel state results, closely matching the ground-truth.}

    \label{novel}
\end{figure*}

The qualitative results on both synthetic and real-world datasets are shown in Figure~\ref{novel}. GaussianEditor (based on inverse rendering) struggles to precisely segment object boundaries, resulting in edge artifacts that necessitate heavy post-processing. Gaussian Grouping (segmentation-feature based) improves performance but still leaves many residual Gaussians, especially for objects with large contact areas between their bottom surface and the background. Graussiancut (graph-based) achieves the best results among the baselines, although slight boundary artifacts remain. DecoupledGaussian incorporates background Gaussian completion, 2D inpainting, and Gaussian fine-tuning. Its 2D inpainting module LaMa produces the most visually coherent results. However, it still struggles to faithfully restore images with complex backgrounds in real-world data. In contrast, our method achieves the highest PSNR and SSIM for novel-state synthesis across both datasets, while maintaining an end-to-end pipeline and avoiding complex multi-stage post-processing.

As shown in Table~\ref{tab:2d_data_synthetic} and~\ref{tab:2d_data_real}, segmentation-only methods yield lower PSNR and SSIM, while adding inpainting improves performance—particularly on synthetic scenes where backgrounds are typically simpler and more structured. In contrast, real-world scenes often involve cluttered or textured backgrounds, making accurate hole filling more challenging and less reliable. Instead of relying on inpainting, we leverage the complementary information across multiple scene states to supervise the optimization of Gaussians, enabling more accurate and consistent scene representation.  
\begin{table}[!t]
 \centering
 \begin{tabular}{l|cc}
    \toprule
    \textbf{Model} & PSNR & SSIM \\
    \midrule
    GaussianEditor (CVPR 2024) & 28.25 & 0.946 \\
    Gaussian Grouping (ECCV 2024) & 28.93 & 0.950 \\
    Gaussiancut (NIPS 2024) & 29.01 & 0.956 \\
    DecoupledGaussian (CVPR 2025) & 30.27 & 0.959 \\
    \textbf{IGFuse (ours)} & \textbf{36.93} & \textbf{0.978} \\
    \bottomrule
 \end{tabular}
 \caption{Quantitative comparison of novel state synthesis quality on the \textbf{synthetic} dataset.}
 \label{tab:2d_data_synthetic}
\end{table}

\begin{table}[!t]
 \centering
 \begin{tabular}{l|cc}
    \toprule
    \textbf{Model} & PSNR & SSIM \\
    \midrule
    GaussianEditor (CVPR 2024) & 21.02 & 0.849 \\
    Gaussian Grouping (ECCV 2024) & 21.68 & 0.853 \\
    Gaussiancut (NIPS 2024) & 21.81 & 0.864 \\
    DecoupledGaussian (CVPR 2025) & 22.28 & 0.855 \\
    \textbf{IGFuse (ours)} & \textbf{27.18} & \textbf{0.907} \\
    \bottomrule
 \end{tabular}
 \caption{Quantitative comparison of novel state synthesis quality on the \textbf{real-world} dataset.}
 \label{tab:2d_data_real}
\end{table}

\subsection{Ablation Study}
\begin{table}[t]
  \centering
  \renewcommand{\arraystretch}{0.9}

  \begin{tabular*}{\columnwidth}{@{\extracolsep{\fill}}ccc|cc}
    \toprule
    B & C & P & PSNR~$\uparrow$ & SSIM~$\uparrow$ \\
    \midrule
    \CheckmarkBold & - & - & 35.10 & 0.971 \\
    \CheckmarkBold & \CheckmarkBold & - & 35.55& 0.974 \\
    \CheckmarkBold & \CheckmarkBold & \CheckmarkBold & \textbf{36.93} & \textbf{0.978} \\
    \bottomrule   
  \end{tabular*}
    \caption{Ablation study of B (Bidirectional alignment), C (Collaborative co-pruning), and P (Pseudo-state guided alignment).}
  \label{tab:ablation}
\end{table}

Table~\ref{tab:ablation} presents the ablation study evaluating the contribution of each component in our framework: Bidirectional alignment (B), Collaborative co-pruning (C), and Pseudo-state guided alignment (P). Using only Bidirectional alignment already provides a strong baseline, achieving a PSNR of 35.10. Introducing co-pruning yields a slight improvement in structural quality. This is because Bidirectional alignment tends to reassign residual Gaussians to have background-like colors or reduced opacity. While co-pruning helps eliminate these floaters, its overall impact on PSNR is limited. In contrast, incorporating Pseudo-state guided alignment results in a substantial increase in PSNR. This improvement arises from the fact that occlusion ambiguities cannot be fully resolved with only two configurations, additional pseudo-states provide richer supervision across multiple viewpoints, enhancing alignment between the two Gaussian fields and leading to more consistent and photorealistic reconstructions.

\begin{table}[t]
  \centering
  \renewcommand{\arraystretch}{0.9}
 
  \begin{tabular*}{\columnwidth}{@{\extracolsep{\fill}}c|c|cc|cc}
    \toprule
     \textbf{Gaussian}&\textbf{Pseudo}& \multicolumn{2}{c|}{\textbf{Synthetic 1}} & \multicolumn{2}{c}{\textbf{Synthetic 2}} \\
     &             & PSNR & SSIM & PSNR & SSIM \\
    \midrule
    $\mathcal{G}_1$   & w/o & 37.04 & 0.979 & 37.51 & 0.977 \\
    $\mathcal{G}_2$   & w/o & 37.16 & 0.977 & 36.14 & 0.974 \\
    \midrule
    $\mathcal{G}_1$ & w/  & 39.26 & 0.984 & 37.59 & 0.977 \\
    $\mathcal{G}_2$ & w/  & 39.26 & 0.984 & 37.50 & 0.977 \\
    \bottomrule
  \end{tabular*}
  \caption{PSNR and SSIM of $\mathcal{G}_1$ and $\mathcal{G}_2$ in Syntheti 1 and Synthetic 2 scenes, with and without pseudo-state supervision.}
  \label{tab:psnr_ssim}
\end{table}

\subsection{Dual Guassian Convergence}

We investigate the convergence behavior of two Gaussian fields trained from different synthetic scenes. In the absence of Pseudo-state Guided Alignment, PSNR and SSIM differ significantly when evaluated in the target state. These discrepancies stem from occlusions and viewpoint differences that lead to misalignments between the two fields. Even with Bidirectional Alignment, such inconsistencies persist, indicating incomplete convergence. By incorporating Pseudo-state guided Alignment, we enforce consistency across object compositions in both fields, allowing them to observe complementary content and provide mutual supervision. This promotes convergence toward a shared and coherent optimized representation. Empirically, Gaussian fields trained from either state yield nearly identical PSNR and SSIM when evaluated under same test configuration, demonstrating effective alignment and mutual consistency.

\subsection{Background Separation}
\begin{figure}[!ht]
    \centering
    \includegraphics[width=1\linewidth]{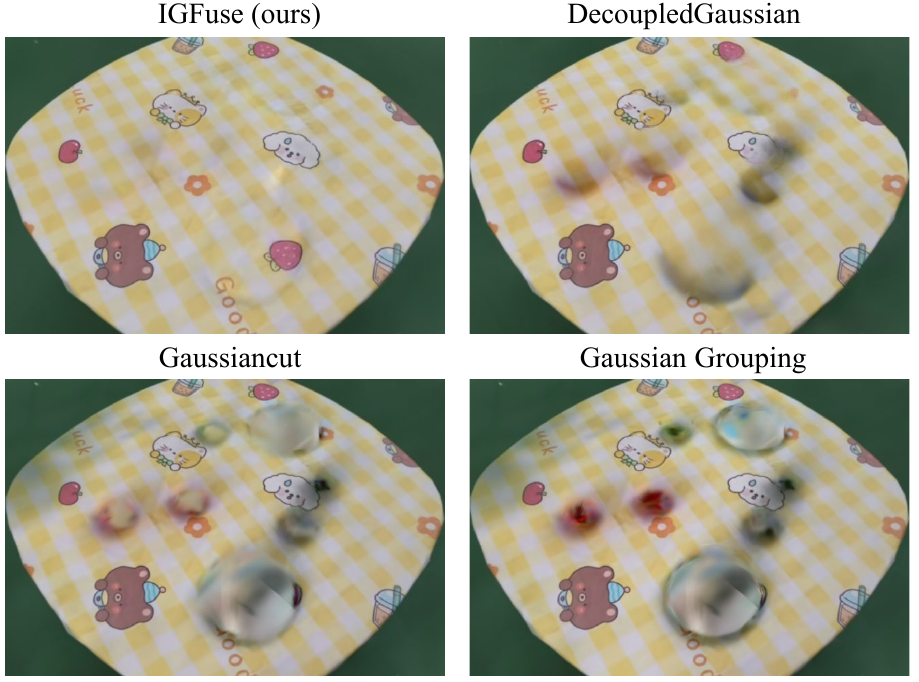}
    \caption{Comparison of different techniques for background separation. IGFuse (ours) vs. Decoupled Gaussian, Gaussiancut, and Gaussian Grouping.}
    \label{bg}
\end{figure}

As shown in Figure~\ref{bg}, when separating only the background, both Gaussian Grouping and Gaussiancut leave residual Gaussians from objects, with larger objects causing noticeable holes. Although DecoupledGaussian employs LaMa to inpaint object mask regions, the inpainting often produces blurry results, especially in complex backgrounds. In contrast, our multi-scan fusion approach effectively generates complete and seamless background reconstructions.

\subsection{Training Iteration}
\begin{figure}[!ht]
    \centering
    \includegraphics[width=1\linewidth]{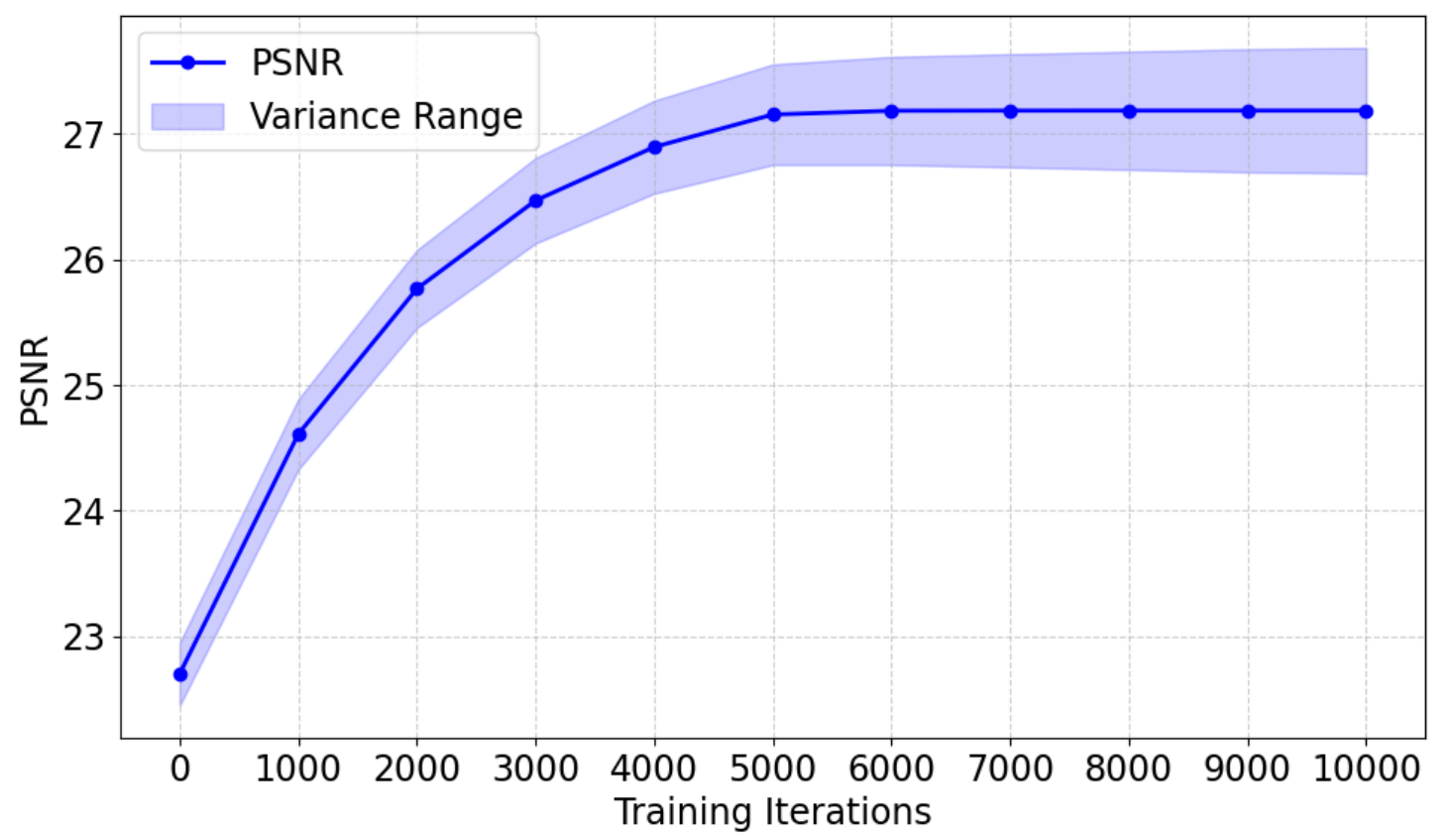}
    \caption{PSNR vs. Training Iterations with Variance Range}
    \label{iter}
\end{figure}
To determine a suitable number of iterations for optimizing the Gaussian fields, we evaluated multiple real-world scenes by measuring the mean and variance of PSNR on the test state across different iteration counts. For each scene with $n$ scans, we normalize the total number of iterations by $n$—that is, the iteration count refers to how many optimization steps each individual scan undergoes. We observe that the PSNR stabilizes around 5000 iterations per scan, indicating convergence. Since we align scene pairs analogously to constructing an undirected graph, we set the final number of training iterations to $n \times 5000$, where $n$ is the total number of scans in the scene. Additionally, we observe a gradual increase in variance. This is because, in early iterations, motion-related artifacts result in uniformly low-quality reconstructions. As optimization progresses and overall quality improves, differences in native PSNR across scenes become more pronounced, leading to increased variance.
\section{Limitations}
\label{limit}
Despite its effectiveness, IGFuse has several limitations. Existing optimization methods are designed for the entire scene. However, since backgrounds across different scans often share similar structures, focusing optimization specifically on object–background boundaries in future work could lead to a more lightweight model. Additionally, our model does not handle lighting variations, causing static shadows even when objects move, which affects realism. Incorporating relighting into the framework could further enhance simulation fidelity in future work.

\section{Conclusion}
\label{conclusion}
We present IGFuse, an end-to-end framework for interactive 3D scene reconstruction via multi-scan fusion. By leveraging object-level transformations across multiple observed scene states, our method overcomes challenges caused by occlusions and segmentation ambiguity. Through bidirectional consistency and pseudo-state alignment, IGFuse refines geometry and semantics to produce high-quality Gaussian fields that support accurate rendering and object-aware manipulation. Extensive experiments validate the effectiveness of our approach for interactive scene reconstruction in vision tasks.

\bibliography{aaai2026}
\appendix
\clearpage
\twocolumn[
\begin{center}
    \LARGE \textbf{IGFuse: Interactive 3D Gaussian Scene Reconstruction via Multi-Scans Fusion}\\
    \vspace{1em}
    \LARGE \textbf{Supplementary Material}
\end{center}
\vspace{5em}
]

\section{Dataset Preparation}
\begin{figure*}[!t]
    \centering
    \includegraphics[width=0.99\linewidth]{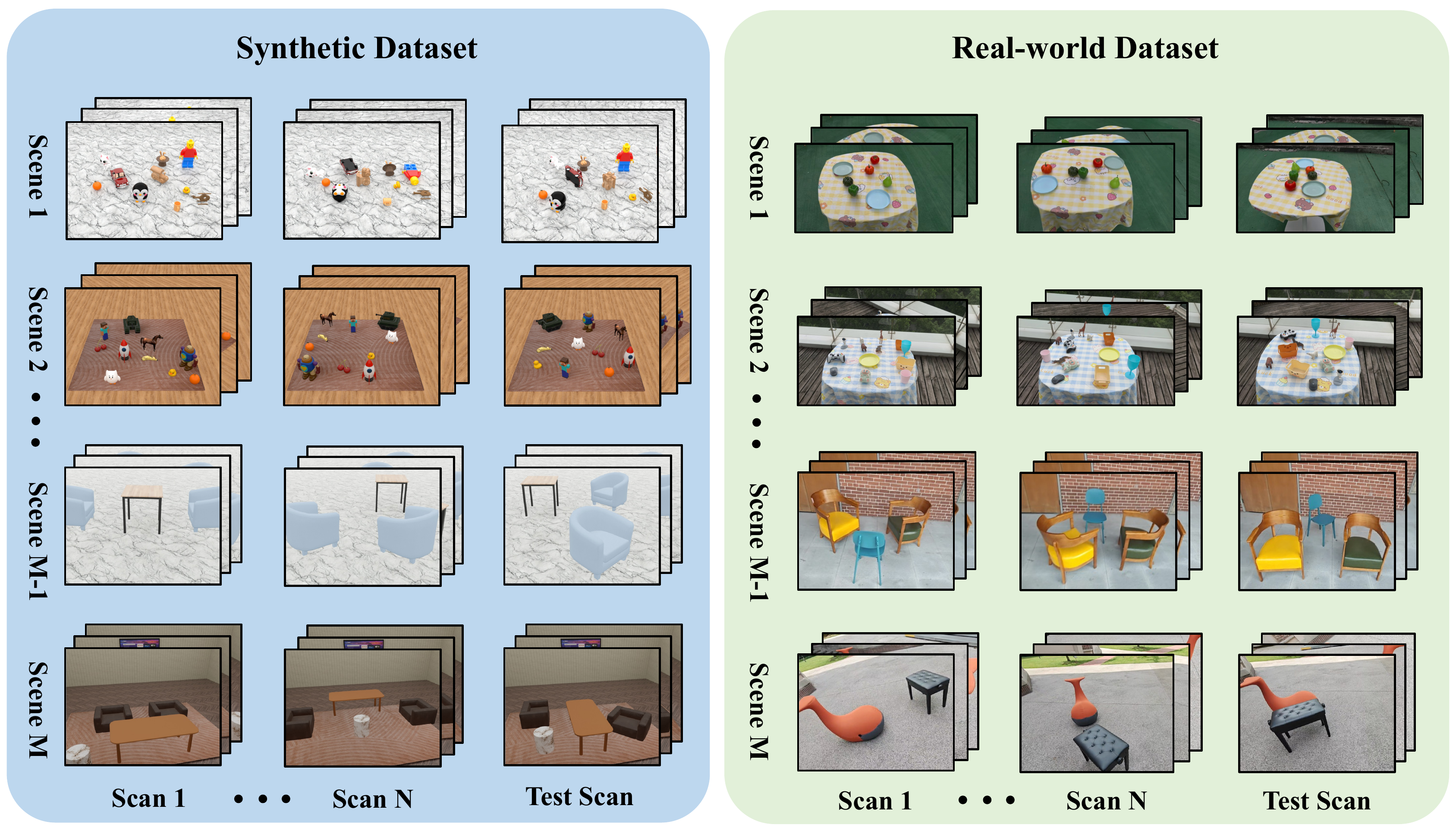}
    \caption{Overview of our dataset, consisting of 7 synthetic scenes (left) and 5 real-world scenes (right). Each scene includes several training scans (Scan 1 to Scan N) with random object layouts and one held-out test Scan for evaluation.}
    \label{data}
\end{figure*}

To support training and evaluation in both synthetic and real-world settings, we construct a multi-scan dataset comprising scenes with multiple static states. In each scene, a subset of objects undergoes random rigid-body perturbations, such as translations and rotations. Our dataset consists of 7 simulated scenes and 5 real-world multi-scan scenes. Each scene includes multiple object configurations, with approximately 100 views captured per scan. In addition to the several training scan data, we also provide a separate test scan data for evaluating reconstruction quality using PSNR and SSIM. An overview of the dataset is illustrated in Figure~\ref{data}.

\subsection{Simulated Scenes}

\begin{figure}[!ht]
    \centering
    \includegraphics[width=1\linewidth]{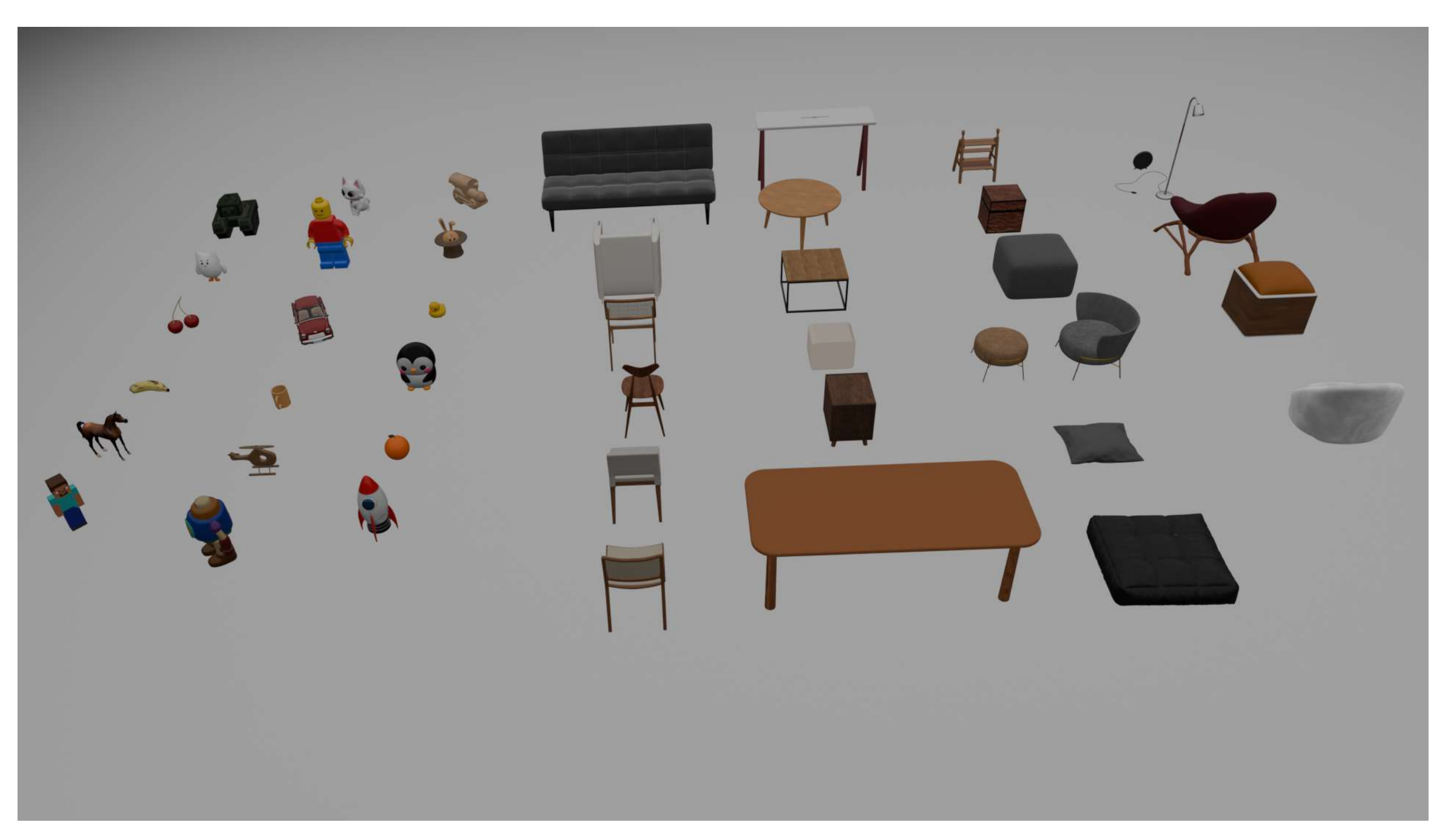}
    \caption{A selection of virtual assets used in our simulated scenes. All objects were sourced from the BlenderKit library~\cite{blenderkit}.}
    \label{sim}
\end{figure}

\paragraph{Object Library and Placement}
We use Blender~\cite{blender} to generate synthetic indoor scenes populated with objects from the BlenderKit library~\cite{blenderkit}, as shown in Figure~\ref{sim}. The objects span various semantic categories such as chairs, tables, toys, and storage items. For each scene, $N=3\sim15$ objects are randomly selected and placed on physically valid surfaces.

\paragraph{Perturbation Strategy}
To generate a number of scans, we apply rigid perturbations for all the objects, including random translations and in-place rotations. The object transformations can be directly obtained from Blender.

\paragraph{Camera Setup}
For each scan, we render $m=100$ perspective views with a resolution of $640\times480$ using uniformly distributed viewpoints on a hemisphere around the scene. Camera intrinsics (focal length, principal point) are shared across both scans to simplify alignment. 

\subsection{Real-World Scenes}

\paragraph{Video Capture}
We record several short video sequences (10–20 seconds) of the same scene using a handheld RGB camera, introducing random changes in object layout between recordings.

\begin{figure}[!ht]
    \centering
    \includegraphics[width=1\linewidth]{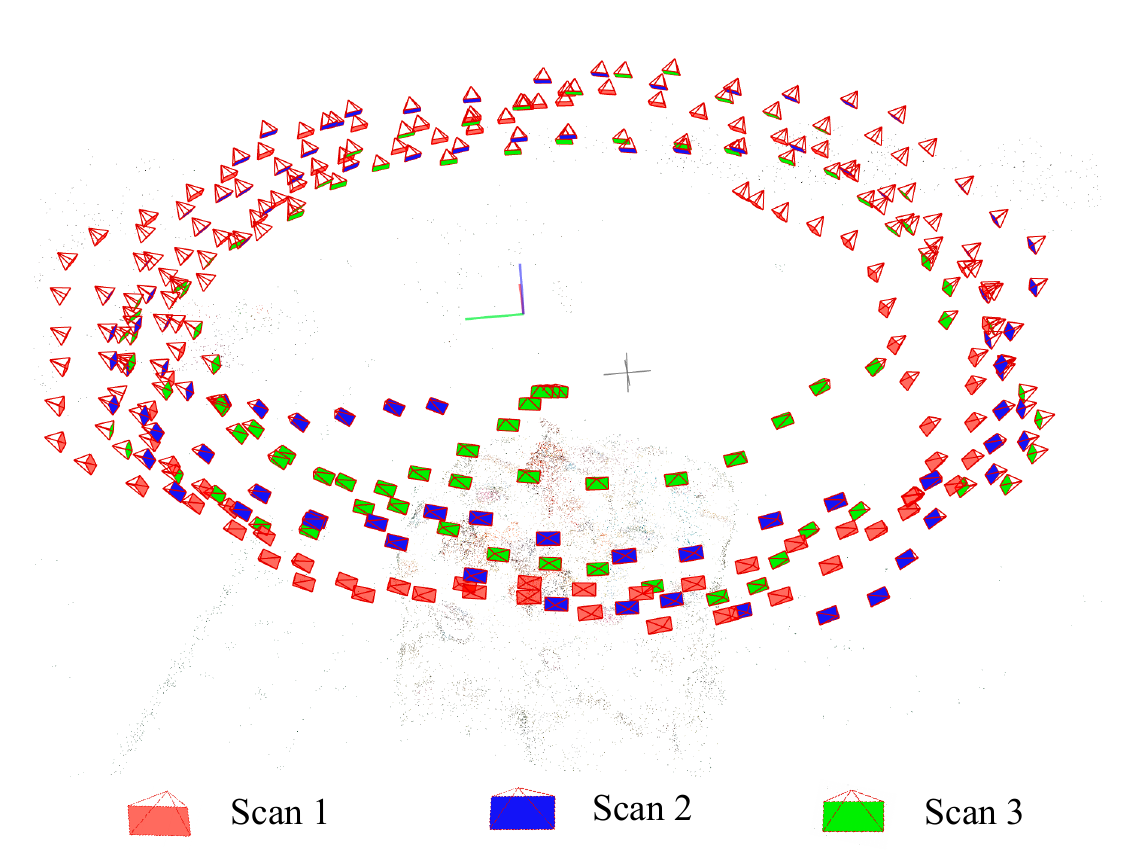}
    \caption{Illustration of the COLMAP used for scene alignment.}
    \label{marker}
\end{figure}
\paragraph{Scene Alignment}

To initialize the point cloud and estimate accurate camera poses, we reconstruct the scene using COLMAP with images from all scans, as shown in Figure~\ref{marker}. Despite variations in foreground objects, COLMAP produces a well-aligned background and reliable poses. This serves as a strong reference for subsequent 3D Gaussian Splatting reconstruction, ensuring consistent background alignment in the final model.

\paragraph{Object Alignment and RT Estimation}
After scene alignment, we extract 3D object masks and compute object-wise transformations between the two scans. Specifically, we first perform per-frame segmentation using SAM2~\cite{ravi2024sam}. For each matched object pair between two scans, an initial transformation matrix is obtained by aligning the object using the Iterative Closest Point (ICP) algorithm. This matrix is then refined by optimizing it with supervision from the object's RGB mask, leading to a more accurate estimation of the transformation.

\paragraph{Segmentation and ID Consistency}
For synthetic data, ground-truth instance masks are rendered per object. For real scenes, segmentation masks are initially generated using SAM2~\cite{ravi2024sam}. To ensure cross-scan consistency, we extract features from images in different scans using DINOv2~\cite{oquab2023dinov2} and compute the cosine similarity between them. Regions with similarity above a predefined threshold are assigned a consistent ID across scans, enabling consistent association across views and scans.

\begin{figure*}[!ht]
    \centering
    \includegraphics[width=1\linewidth]{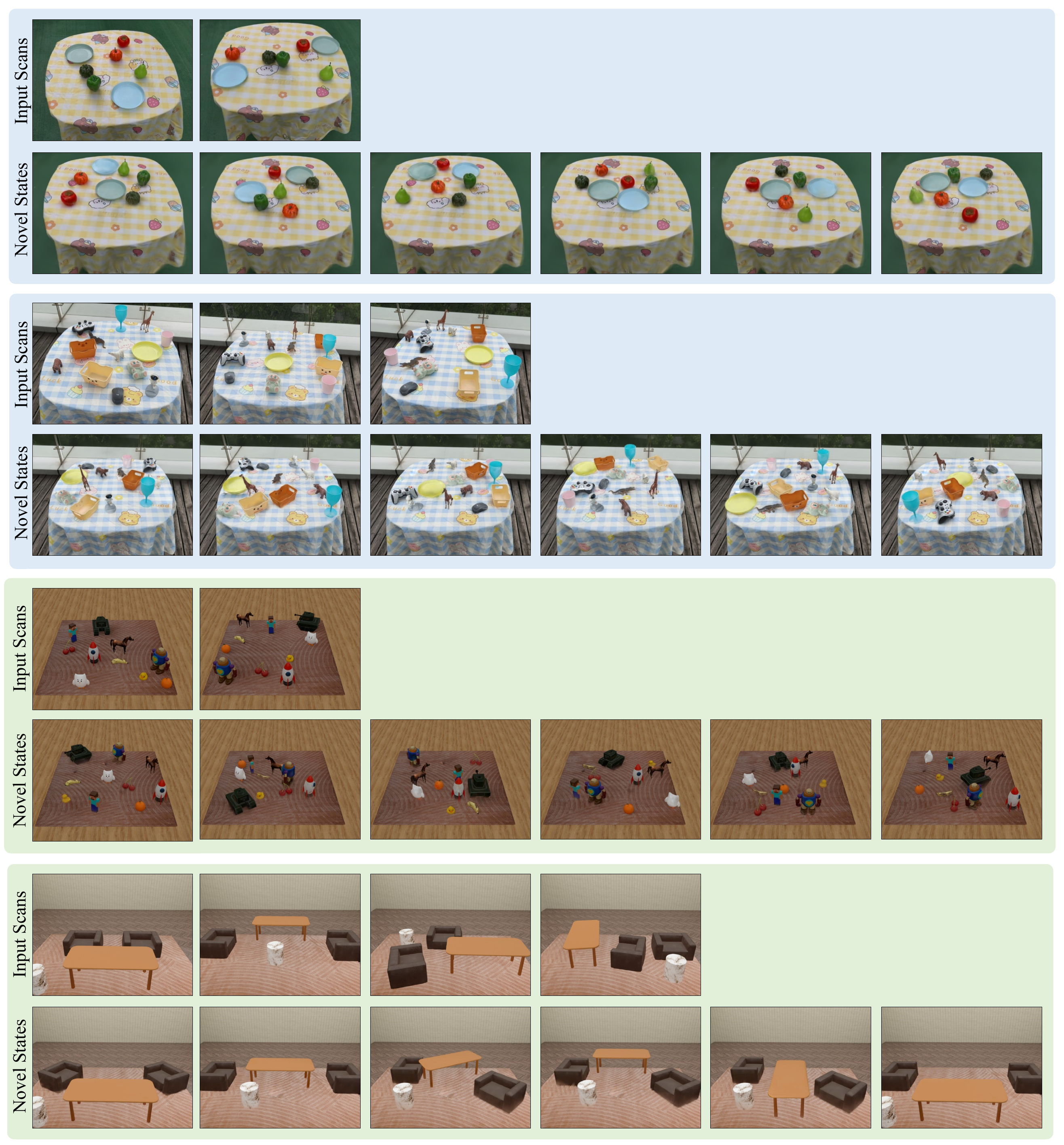}
    \caption{Novel state synthesis results. For each scene, the top row shows the object arrangements in the input scans, while the bottom row presents the synthesized novel states under randomly arranged object configurations.}
    \label{novel states}
\end{figure*}

\section{Pseudo State Construction}
To facilitate dual-state alignment, we construct a pseudo state using boundary detection and object overlap avoidance algorithms. First, we analyze the object positions across all scans to determine a shared scene boundary. Then, we compute the minimum object spacing threshold by averaging the three smallest inter-object distances. Objects are then sequentially placed into the scene using random rotations and translations. For each placement, we check whether the object is too close to any previously placed object—if the distance falls below the threshold, a new position is sampled. This process repeats until all objects are successfully placed. Finally, we perform a bounding box check based on each object’s point cloud to ensure no geometric overlap occurs.

\section{Novel State Synthesis}
In Figure~\ref{novel states}, we show the input scans alongside the resulting novel states. As illustrated, our method supports the generation of novel states with arbitrary object arrangements while maintaining high visual quality. It is also evident that leveraging just two input scans already leads to noticeable improvements in the reconstruction results.
\begin{figure*}[!t]
    \centering
    \includegraphics[width=1\linewidth]{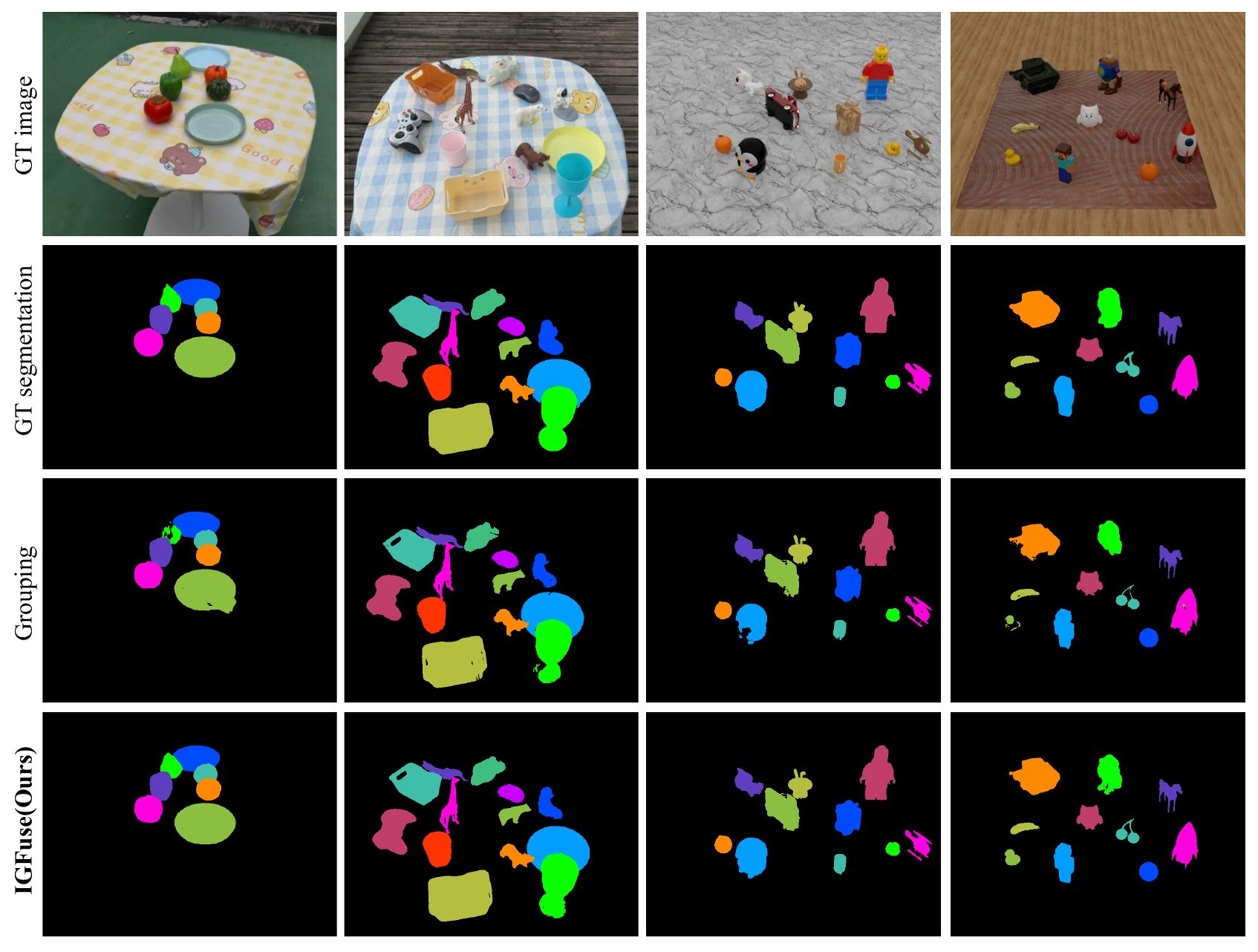}
    \caption{Qualitative comparison of segmentation results across different scenes. From top to bottom: ground-truth (GT) image, ground-truth segmentation, results of Gaussian Grouping, and our method IGFuse. Our method produces more accurate and consistent segmentations, especially around object boundaries, closely matching the ground truth.}

    \label{seg}
\end{figure*}

\section{Details of $\mathcal{L}_\text{r}$}
As referenced in the main text, we elaborate here on the formulation of the overall loss $\mathcal{L}_\text{r}$~\cite{ye2023gaussian}. We use a linear layer $f$ followed by a softmax activation to map the rendered 2D features $\textit{S}$ into a $\mathcal{K}$-class semantic space. For the resulting 2D classification, we apply a standard cross-entropy loss $\mathcal{L}_{2d}$ to enforce accurate pixel-level mask predictions:
\begin{equation}
    \mathcal{L}_\text{2d} = - \sum_{k \in \mathcal{K}} \boldsymbol{m}[k] \log \left(\text{softmax}(f(\textit{S}))[k]\right)
\end{equation}

To ensure consistency in 3D segmentation, we further introduce a 3D segmentation loss $\mathcal{L}_{3d}$. This is a cross-entropy loss applied to the per-point segmentation features $s$, using the fused 3D pointmap labels $\mathcal{P}^{s}$ as pseudo ground truth:
\begin{equation}
    \mathcal{L}_\text{3d} = - \sum_{k \in \mathcal{K}} \mathcal{P}^{s}[k] \log \left(\text{softmax}(f(\textit{s}))[k]\right)
\end{equation}

In addition, we adopt the standard 3D Gaussian image reconstruction loss as proposed in~\cite{kerbl20233d}, which is a weighted combination of $L_1$ loss and D-SSIM loss:
\begin{equation}
    \mathcal{L}_\text{img} = (1 - \lambda) \mathcal{L}_1 + \lambda \mathcal{L}_\text{D-SSIM}
\end{equation}

The final training objective $\mathcal{L}_\text{r}$ is a weighted sum of all three components:
\begin{equation}
    \mathcal{L}_\text{r} = \mathcal{L}_\text{img}
    + \lambda_\text{2d} \mathcal{L}_\text{2d}
    + \lambda_\text{3d} \mathcal{L}_\text{3d}
\end{equation}
In our experiments, we set $\lambda_\text{2d} = 1$ and $\lambda_\text{3d} = 1$.
\begin{figure*}[!t]
    \centering
    \includegraphics[width=1\linewidth]{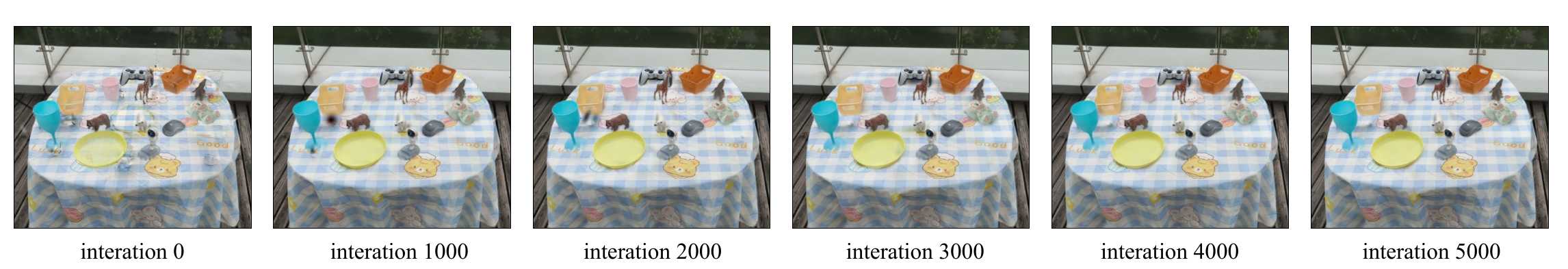}
    \caption{Visualizations result over different training iterations.}
    \label{iter_fig}
\end{figure*}

\begin{figure*}[!t]
    \centering
    \includegraphics[width=1\linewidth]{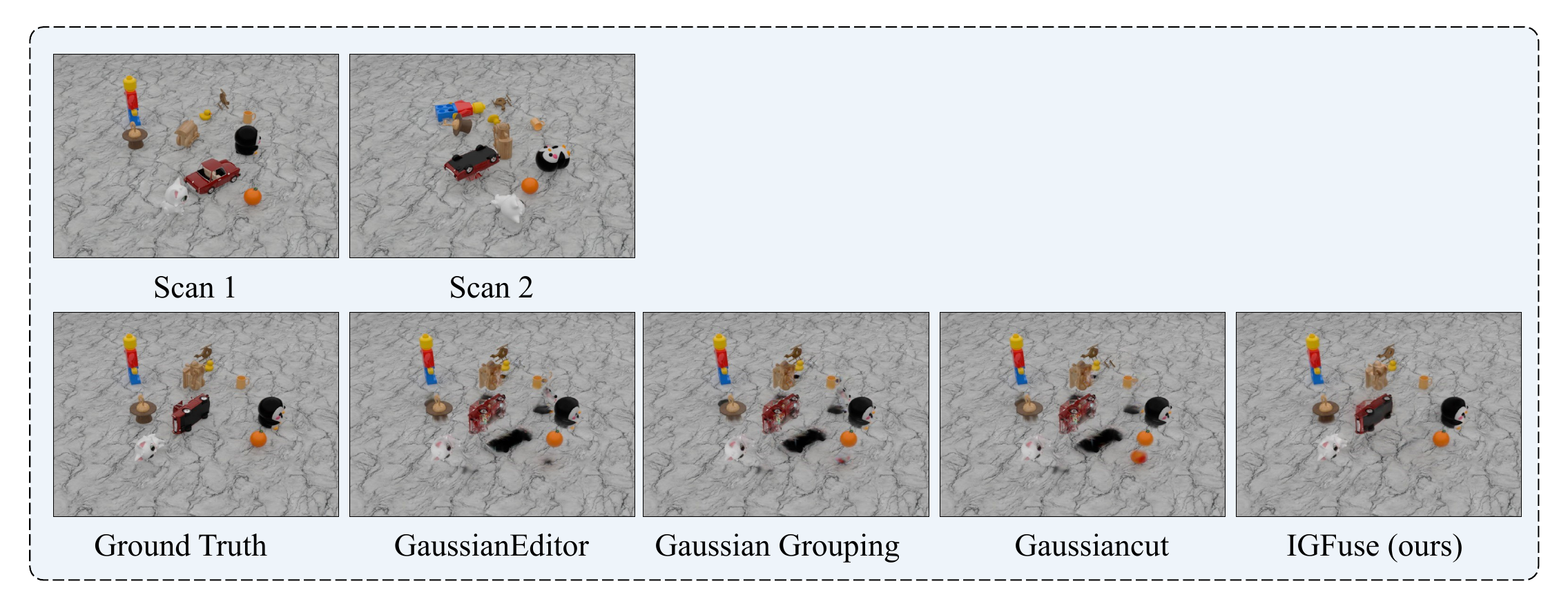}
    \caption{Multi-scan object observation results. Our method effectively fuses observations from multiple scans to reconstruct complete object models.}
    \label{fall}
\end{figure*}
\section{Segmentation Improvement}
As shown in Figure~\ref{seg}, our multi-scan segmentation method provides improvements over the feature-based baseline, Gaussian Grouping, particularly in fine-grained details. Gaussian Grouping often produces artifacts around object boundaries, while our results are much closer to the ground truth. Quantitatively, the mIoU improves by approximately 4\%, as shown in the Table~\ref{miou}.

\begin{table}[h]
\centering
\begin{tabular}{@{}lc@{}}
\toprule
\textbf{Method} & \textbf{mIoU (\%)} \\
\midrule
Gaussian Grouping     & 86.8 \\
IGFuse (Ours)   & \textbf{91.0} \\
\bottomrule
\end{tabular}
\caption{Comparison of mIoU for different methods}
\label{miou}
\end{table}

\section{Training Iteration}
We provide visualizations in Figure~\ref{iter_fig} for training iteration. At iteration 0, when Bidirectional Alignment and Pseudo-state Guided Alignment are not yet applied, many artifacts remain in the scene due to inaccurate object transitions. As optimization proceeds, these artifacts are gradually eliminated. By iteration 5000, the optimization is nearly complete, and the artifacts have mostly disappeared.

\section{Multi-scans Object Observation}
In the main text, we demonstrate how multi-scan fusion enhances the completeness of the background. This benefit also extends to object-level reconstruction, as illustrated in Figure~\ref{fall}. In scan 1 and scan 2, only the top and bottom of the small red car are visible, respectively, making it difficult to obtain a complete observation of the object. In contrast, the second row shows that other methods based on a single scan (e.g., scan 1) generate novel states where the bottom of the car is severely missing. By fusing information from both scan 1 and scan 2, our method reconstructs a complete object model and supports arbitrary object configurations in novel scenes.

\begin{figure}[!ht]
    \centering
    \includegraphics[width=1\linewidth]{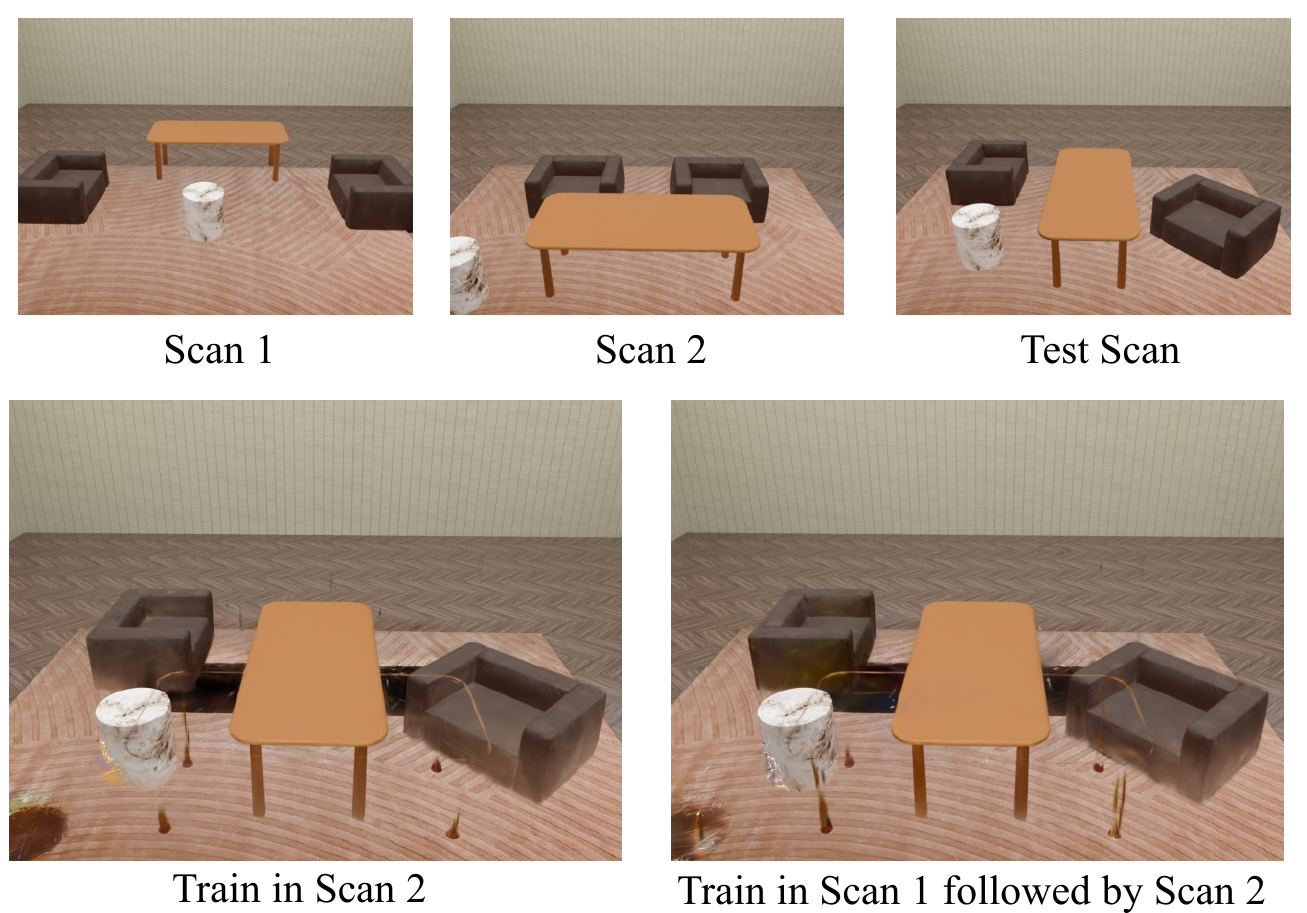}
    \caption{Comparison of training strategies using different scene scans. The top row shows three static observations: Scan 1, Scan 2, and a Test Scan. The bottom row presents the rendering results at the Test Scan after training on Scan 2 only (left) and after sequential training on Scan 1 followed by Scan 2 (right). The results show that incorporating an additional scan does not bring significant improvement over single-Scan training.}
    \label{fair}
\end{figure}

\begin{table}[ht]
  \centering
  \begin{tabular}{l|l|c|c}
    \toprule
    \textbf{Training data} & \textbf{Model} & \textbf{PSNR} & \textbf{SSIM} \\
    \midrule
    \multirow{2}{*}{Single-scan}&
    Gaussian Grouping & 28.93 & 0.950\\
    &DecoupledGaussian &30.27 & 0.959 \\
    \midrule
    \multirow{2}{*}{Multi-scans}&
     Gaussian Grouping & 28.99& 0.952 \\
    & DecoupledGaussian & 30.29 &0.959 \\
    \midrule
    Multi-scans & IGFuse (Ours) & \textbf{36.93} & \textbf{0.978} \\
    \bottomrule
  \end{tabular}
   \caption{Quantitative comparison of 3D Gaussian segmentation methods with single-scan and multi-scans training. Multi-scans inputs yield slight gains, while inpainting dominates performance. IGFuse achieves the best results without relying on inpainting.}
    \label{tab:fair}
\end{table}
\section{Fair Comparison with Single-Scan Methods under Multi-Scans Training}
Since multi-scan input provides more information than single-scan settings, direct comparisons may be unfair. To ensure fair evaluation, we adapt single-scan segmentation methods to operate under the multi-scan setting as well. Specifically, we use two scans: the model is first trained on scan 1 to obtain segmentation results, and then the segmented objects are transferred to scan 2, where training continues under supervision from the images in scan 2. This approach produces results that are nearly identical to those obtained by training directly on scan 2. The unoccluded ground observed in scan 1 is not preserved after the transfer. Instead, it becomes missing regions, as shown in Figure~\ref{fair}. This demonstrates a key limitation of single-scan methods in fusing information across scans, while our Bidirectional Alignment strategy effectively addresses this issue.

As shown in the Table~\ref{tab:fair}, incorporating multi-scans brings only slight improvements in PSNR and SSIM for methods like Gaussian Grouping. This minor gain is attributed to small residual Gaussians that may remain on the ground under certain objects, as seen in the bottom left of Figure~\ref{fair}. However, after inpainting is introduced, the performance of DecoupledGaussian is nearly identical, since the missing regions are mainly filled by the inpainting process and the additional scan training provides little further benefit.

\section{Depth Consistency Comparison}

\begin{figure}[!ht]
    \centering
    \includegraphics[width=1\linewidth]{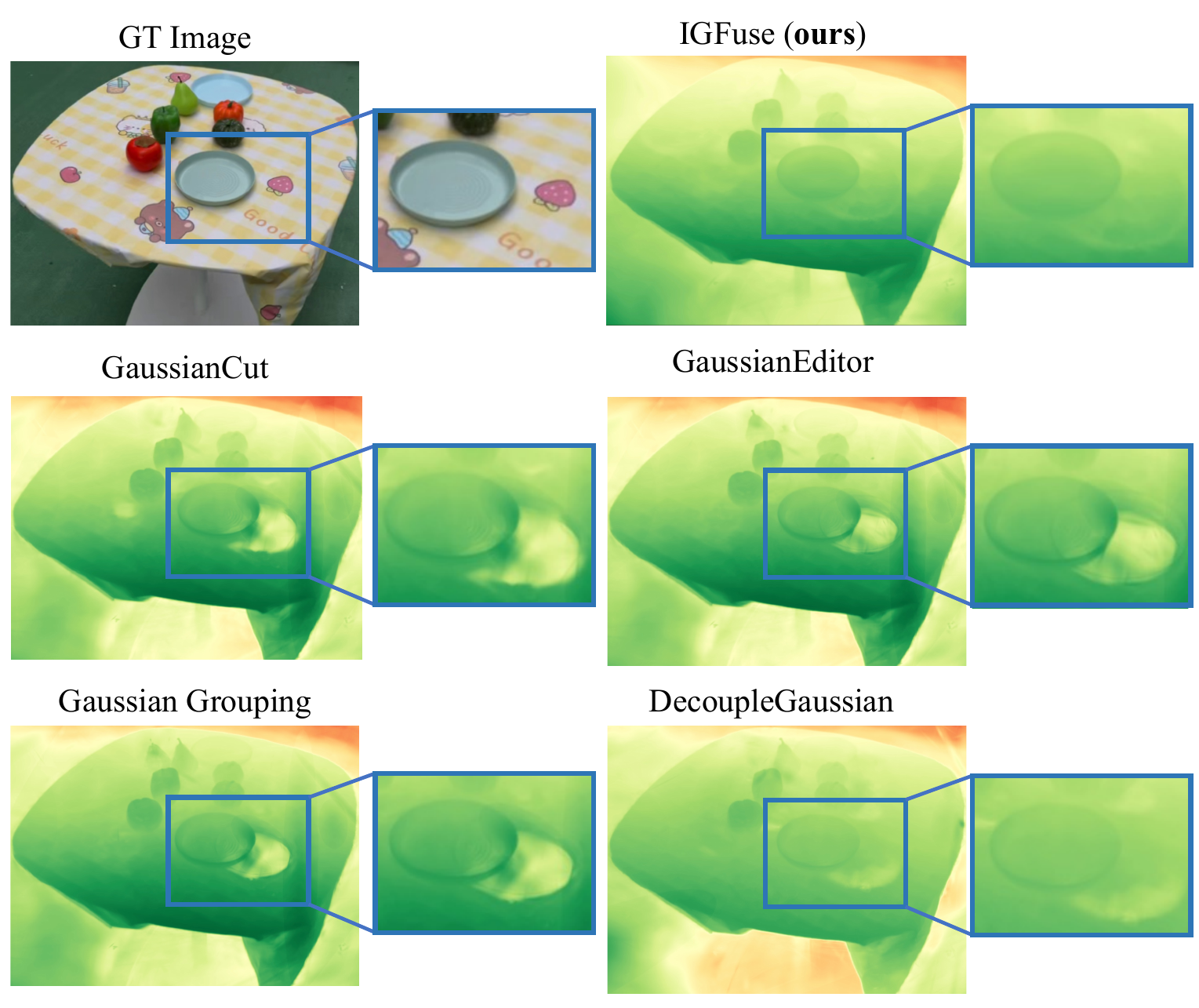}
    \caption{Depth consistency comparison across different methods.}
    \label{depth}
\end{figure}

Although no depth or normal supervision is used and only the original 3D Gaussians are employed, our method achieves superior depth reconstruction due to the effective supervision from multi-scan scenes. Other methods suffer from visible holes after removal of the object, as shown in Figure~\ref{depth}. Although the inpainting-based DecoupledGaussian method performs better in filling these gaps, some depth inconsistencies still remain due to the separate optimization of foreground and background.

\end{document}